\documentclass[letterpaper, 10 pt, conference]{ieeeconf}  

\IEEEoverridecommandlockouts                              

\usepackage{url}      
\usepackage[]{graphicx}
\usepackage[]{tabularx}
\graphicspath{{03_graphics/}}
\usepackage{amsmath}
\usepackage{amssymb}
\usepackage{algorithm}
\usepackage{algpseudocode}
\usepackage{siunitx}
\usepackage{xcolor,color,soul}
\usepackage{definitions}
\usepackage{macros}
\usepackage[absolute,overlay]{textpos}

\usepackage[colorinlistoftodos, english, disable]{todonotes} 

\usepackage[utf8]{inputenc}

\usepackage[font=footnotesize]{caption}
\usepackage[font=footnotesize]{subcaption}

\usepackage{hyperref}

\title{\LARGE \bf Accelerating Cooperative Planning for Automated Vehicles with Learned Heuristics and Monte Carlo Tree Search}

\author{Karl Kurzer$^{1}$, Marcus Fechner$^{1}$ and J. Marius Z\"ollner$^{1,2}$ 
	\thanks{
		$^{1}$Karlsruhe Institute of Technology, Kaiserstr. 12, 76131 Karlsruhe, Germany
		{\tt\small kurzer@kit.edu}
		$^{2}$FZI Research Center for Information Technology, Haid-und-Neu-Str. 10-14, 76131 Karlsruhe, Germany
		{\tt\small fechner@fzi.de, zoellner@fzi.de}
	}
}        
        
\begin{document}
\begin{textblock*}{\textwidth}(19mm,10mm)
	\footnotesize
	\noindent\textcopyright2020 IEEE. Personal use of this material is permitted.  Permission from IEEE must be obtained for all other uses, in any current or future media, including reprinting/republishing this material for advertising or promotional purposes, creating new collective works, for resale or redistribution to servers or lists, or reuse of any copyrighted component of this work in other works.\\
	\textit{2020 IEEE Intelligent Vehicles Symposium (IV)}
\end{textblock*}	
\listoftodos

\maketitle
\thispagestyle{empty}
\pagestyle{empty}

\begin{abstract}
Efficient driving in urban traffic scenarios requires foresight.
The observation of other traffic participants and the inference of their possible next 
actions depending on the own action is considered cooperative prediction and planning.
Humans are well equipped with the capability to predict the actions of multiple interacting traffic participants and plan accordingly, without the need to directly communicate with others.
Prior work has shown that it is possible to achieve effective cooperative planning without
the need for explicit communication.
However, the search space for cooperative plans is so large that most of 
the computational budget is spent on exploring the search space in unpromising regions that are far 
away from the solution.
To accelerate the planning process, we combined learned heuristics with a cooperative planning method to guide the search towards regions with promising actions,
yielding better solutions at lower computational costs.
\end{abstract}

\section{Introduction}\label{sec:introduction}
Cooperative planning methods consider the mutual dependence of actions in multi-agent environments,
opposed to methods that reduce multi-agent to single-agent environments,
with other agents' actions being independent of one another.
This reduction accelerates the planning process heavily,
as it addresses the curse of dimensionality.
But, the lack of consideration prunes the solution space,
so that solutions requiring cooperation of agents can not be discovered.
Thus, more suitable methods need to be developed, in order to handle the complexity of this problem class without dropping the interdependence of actions.

While a chess player could consider all movable pieces before taking a decision,
even novice players quickly exclude certain moves that they deem irrelevant given the current board state.
It is the ability to combine search with learned heuristics that allows us to discover solutions for sequential decision making problems quickly instead of being stuck thinking \cite{Kahneman2003, Anthony2017}.

Problems solved by iterative methods, converging to a (local) optima, greatly benefit from an
initialization that is as close as possible to the solution. 
Humans use their problem specific experience as the initialization.
Thus, it is not surprising that the super human performance that powered AlphaGo, a computer program that beat the strongest Go player in the world in 2016, results from this combination of search and learned heuristics \cite{Silver2016}.

\begin{figure}
	\centering
	\def\svgwidth{\columnwidth}
	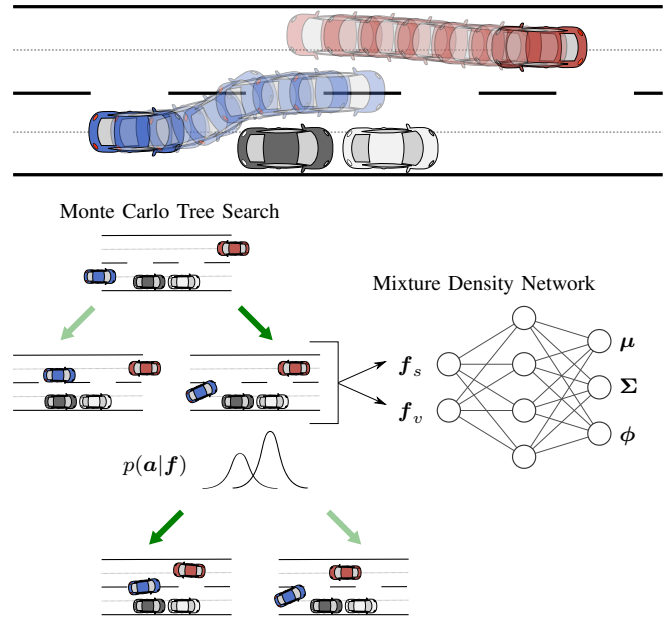
	\caption{Integration of a Mixture Density Network into Monte Carlo Tree Search;
		During the expansion phase (i.e. the exploration of the actions space),
		the current state of the Monte Carlo Tree Search (MCTS) is being transformed into a feature vector $\features$, consisting of scalar features $\featuresscalar$ as well as visual features $\featuresvisual$.
		These features are fed into an mixture density network (MDN), generating Gaussian mixture models (GMMs) from which actions $\action$ are sampled for each agent, biasing the expansion towards auspicious future states. Figure adapted from \cite{Kurzer2018}.
	}
	\label{fig:MCTSMDN}
\end{figure}

The interplay of learned models and search has additional inherent advantages.
Even though the goal for a learned model is to generalize well, i.e. it should perform equally well on known and on unknown data,
it cannot be predicted, what happens when it is fed with data originating from a different distribution.
Furthermore, the lack of introspection makes it hard to encode constraints in the model that guarantee safety.
However, these inherent disadvantages can easily be addressed by search algorithms, as they are bias free and allow for the integration of constraints.

In the domain of automated driving, cooperative multi-agent trajectory planning tasks are costly problems requiring further research.
From a game-theoretic point of view, it is the goal to find a Nash equilibrium, i.e. no single agent can perform a different action yielding a higher return.
Even though cooperative multi-agent trajectory planning algorithms have shown promising results \cite{Kurzer2018,Kurzer2018a, Bahram2016, Lawitzky2013}, their convergence speed is slow compared to state of the art single-agent trajectory planning methods.

The contributions of our work are twofold.
First, we developed a compact and accurate model, using a deep neural network (DNN), which is capable of predicting sampling distributions over actions for cooperative urban multi-agent scenarios.
Second, we integrated the trained model as a heuristic in the cooperative multi-agent trajectory planning algorithm presented in \cite{Kurzer2018a} and evaluated its performance on a variety of different scenarios, see Fig. \ref{fig:MCTSMDN}. 

This paper is structured as follows:
First, a brief overview of research on prediction and learned heuristics used in search algorithms is given in section \ref{sec:relatedWork}.
Section \ref{sec:problemStatement} defines the problem.
The general approach to the problem is presented in section \ref{sec:approach}.
At last, a detailed comparison between the baseline and the proposed versions of the cooperative multi-agent trajectory planning algorithm is conducted in section \ref{sec:evaluation}.

\section{Related Work}\label{sec:relatedWork}

\subsection{Prediction}
The purpose of prediction is to model future events.
For a traffic situation with multiple traffic participants, this entails to estimate where each of the traffic participants will be at a given future time step.
Prediction is the fundamental basis for single-agent planning, as it provides the constraints for the consecutive planning step.

Approaching prediction from a multi-agent perspective, it becomes clear that prediction and planning are non-separable \cite{Bahram2016}. The most likely prediction must incorporate the plan for each agent and thus, the plan for the predicting traffic participant.

However, for reasons of clarity, the term prediction shall denote output of a system, that is not being fed directly into a controller, but is rather incorporated in a subsequent planning step.

Previous work for interactive scene prediction uses DNNs to predict the action of an ego vehicle in merge scenarios for highway driving \cite{Lenz2017}.
Based on features, such as spatial relation, longitudinal time gap of the seven closest vehicles, the road
geometry, and data about the ego vehicle's state, a DNN is trained to parameterize a Gaussian mixture model estimating the longitudinal acceleration and the lateral velocity of the ego vehicle.

Similarly, with an LSTM encoder-decoder architecture with convolutional social pooling, trajectories can be predicted for six distinguished maneuver classes, based on the spatial configuration of a highway traffic scene, capturing the interdependence between vehicles \cite{Deo2018a}.

Other approaches make use of multi-agent trajectory homotopy, enumerating possible maneuver classes using a formation tree \cite{Schulz2017}.
These classes can then be used to estimate the likelihood for any of the maneuver classes being picked given a set of observed trajectories, using Bayesian statistics.
This approach allows for local rather than global estimation of the trajectory within a given homotopy class.
Tight maneuvers such as passing a point at the same time cannot be captured and hence, cannot be predicted.

Using a predefined set of maneuvers for each agent in a highway scenario, the collision probabilities for all permutations of these sets can be computed.
This allows a prediction for the most likely maneuver combination with a look ahead of one time step \cite{Lawitzky2013}.
Due to the short look ahead in combination with a fixed number of maneuvers, cooperative plans in urban driving scenarios
cannot be predicted correctly.
While Bahram et al. \cite{Bahram2016} extend the approach to longer planning horizons, it does not incorporate continuous action spaces, limiting its applicability to highway driving.

Additionally, stochastic approaches, based on dynamic Bayesian models, that can learn conditional dependencies for interactive traffic models from observation using Expectation-Maximization have been proposed \cite{Gindele2015}.
In combination with random forests, a policy modeled as a conditional density function is approximated, predicting future actions.

\subsection{Learned Heuristics}
Knowing about the capabilities as well as the limitations of learned models, the integration of these models into classical planning techniques has been an active field of research.

Silver et al. demonstrated the capabilities of Monte Carlo Tree Search (MCTS) in combination with learned value and policy networks, outranking the state of the art Go programs by a large margin.
They used a policy network to predict priors for each of the possible actions of a node.
These prior probabilities guide the search towards promising areas of the search space decaying over time \cite{Silver2016}.
Similar methods improving the strength of MCTS were developed, interweaving policy learning and roll-outs of the MCTS.
In this case, the MCTS recursively improves itself using the knowledge gathered from previous simulations embedded in a sampling policy that then improves roll-outs further \cite{Anthony2017, Silver2017, Schrittwieser2019}.

In the autonomous driving domain, Hubschneider et al. combine an end-to-end trained DNN, proposing trajectories to a particle swarm optimizer (PSO) \cite{Hubschneider2017}.
The DNN is trained via imitation learning, using visual input from a front-facing
camera and steering angle labels generated from an expert driver.
Using the trained model as a heuristic provided for the initialization of the PSO planner,
the optimization is sped up.
This is especially helpful in scenarios with many static obstacles or tight passages,
where the majority of particles are in a colliding state, since collision checking is a major
bottleneck of any motion planning algorithm \cite{Ziegler2010}.

Related work by Paxton et al. integrate a learned high-level options policy in combination with a low-level control policy into the MCTS to improve the overall quality for a single agent planning problem.
\cite{Paxton2017b}

Other approaches to accelerate sampling based motion planning task have been proposed, using conditional variational autoencoder, generating subspaces for sampling distributions over desired states \cite{Ichter2017}.
Similarly, Banzhaf et al. learn a sampling distribution for poses of an RRT* path planning algorithm in semi-structured environments.
The model is trained via supervised learning, using an occupancy grid representation, enriched with additional features,
such as the start position, the goal position, and the past path.
The output of the model is the predicted path and heading of the vehicle.
Using a sampling algorithm, proposals are drawn from the prediction \cite{Banzhaf2018}.


%
%

While previous work in the area of prediction and integration of learned heuristics exists,
our work demonstrates the feasibility to efficiently predict distributions of trajectories for multiple agents in a single forward pass.
This prediction is then integrated as a heuristic to improve cooperative multi-agent trajectory planning in continuous spaces for urban driving scenarios.

\section{Problem Statement}\label{sec:problemStatement}
\begin{figure}
	\fns
	\centering
	\def\svgwidth{\columnwidth}
	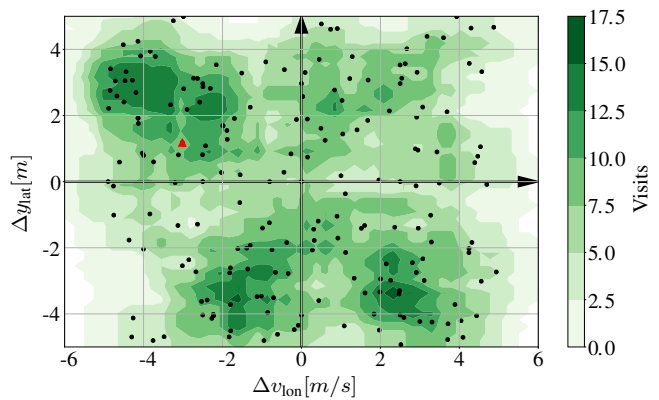
	\caption{Exploration of the unbiased algorithm; darker areas represent regions with high visit counts, meaning that the algorithm has more thoroughly explored these regions. The red triangle is the action that is finally selected. Figure adapted from \cite{Kurzer2018a}}
	\label{fig:ExplorationAnalysis}
\end{figure}

With the goal to guide sampling based, cooperative, multi-agent trajectory planning algorithms, such as \cite{Kurzer2018a} towards promising regions of the action space, a mapping from the feature space 
of the scene to the action space of each agent in the scene needs to be defined, $\featurespace\to\actionspacei$.
Fig. \ref{fig:ExplorationAnalysis} depicts the exploration of a single agent's action space (change in velocity $\Delta v_\text{lon}$ and change in lateral position $\Delta y_\text{lat}$) for the unbiased algorithm developed in our previous work \cite{Kurzer2018a}.
Samples are distributed in a rather random fashion with some regions yielding higher visit counts.
While exploration is necessary in order to find the global optimum, it hinders the exploitation and, therefore, the thorough evaluation of areas of the action space that have shown high returns on some trajectories.
A heuristic that generates actions or allows to derive actions for these kind of problems is thus likely to yield better results at a lower computational cost.
Designing a heuristic based on expert knowledge for planning cooperative trajectories is not only time intensive, but prone to errors, due to the complexity which arises from the interdependence and rare edge cases.
Hence, we decided to learn this heuristic from data.
With the goal to use it in a sample based planner for cooperative driving, the generation of the sampling distribution must meet a compromise between speed and accuracy, so that it can be executed sufficiently often, while ensuring no misguidance.

%
%

\section{Approach}\label{sec:approach}
Since the heuristic is learned from experience, we employed a DNN for function approximation,
mapping the feature space $\featurespace$ of a multi-agent traffic scene with up to eight traffic participants to a distribution over the action space $\actionspacei$ for each agent.

We defined the outputs of the approximator as the parameters of the sampling distribution, namely a Gaussian mixture model (GMM).
This is convenient for three reasons:
Firstly, Gaussian mixtures allow for easy sampling, secondly, they require little parameterization, facilitating the learning as well as reducing the complexity of the output, and lastly, are able to approximate arbitrary probability density functions.

A Gaussian mixture model is a linear superposition of Gaussians.
The probability of a sample $x$ belonging to that distribution is defined by \eqref{eq:GaussianMixture}.

\begin{equation}\label{eq:GaussianMixture}
	p(x) = \sum_{k=1}^{K}{\phi_k\mathcal{N}(x|\mu_k,\Sigma_k)}
\end{equation}
with $K$ being the number of mixture components, $\mu_k$ the mean, $\Sigma_k$ the covariance and $\phi_k$ the mixing coefficient of the respective component.
In order to accommodate for scenarios, where the distribution over the action space is multimodal, e.g. two or more homotopy classes exist (passing and object on the right or left, merging in front or behind), we chose to train a Gaussian mixture model with two and three components to approximate the sampling distribution.

\subsection{Hybrid Model Architecture}
With the goal to approximate a GMM, the outputs of our model estimate the required parameters for the GMM.
These so called mixture density models (MDN) have first been conceived by Bishop \cite{Bishop1994}, with the GMM being conditioned on the feature vector $\features$, see \ref{eq:MDN}.

\begin{equation}\label{eq:MDN}
p(x|\features) = \sum_{k=1}^{K}{\phi_k(\features)\mathcal{N}(x|\mu_k(\features),\Sigma_k(\features))}
\end{equation}

\begin{figure*}
	\centering
	\def\svgwidth{\textwidth}
	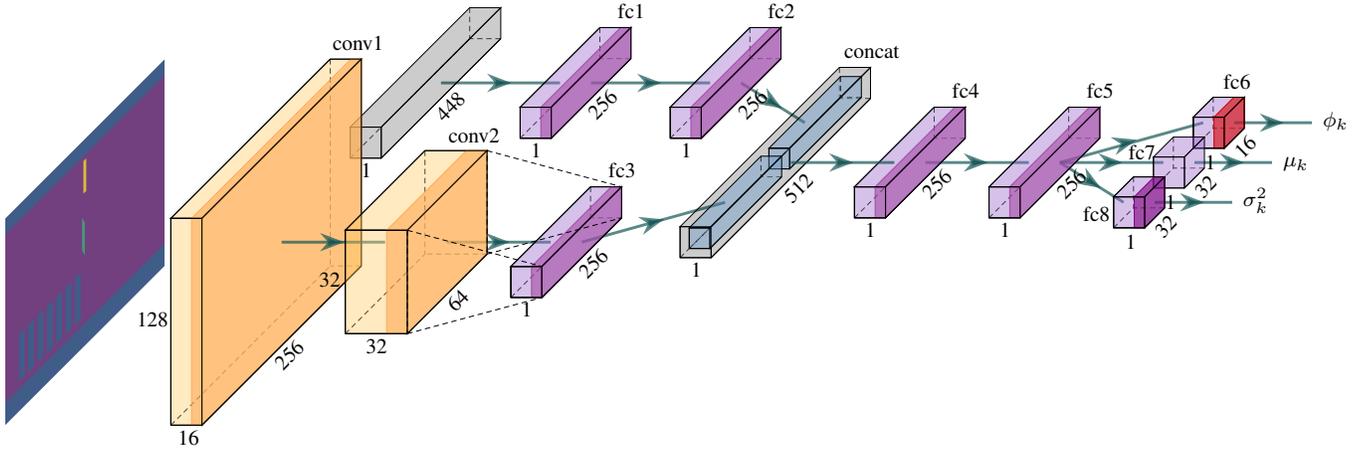
	\caption{Hybrid Model Architecture of the MDN; fully connected layers and convolution layers are denoted by \textit{fc} and \textit{conv} respectively. The network is split into a visual as well as scalar pipeline, handling the respective features. Scalar features of up to eight agents over eight time steps (heading, position, velocity, acceleration, as well as desired velocity and desired lane) are fed into fc1(ReLU) and further processed by fc2(ReLU). Visual features are first processed by conv1(ReLU, filters = 16, kernel = 7x7, stride = 4, padding = reflect) and further fed into conv2(ReLU, filters = 32, kernel = 3x3, stride = 1, padding = reflect). After this they are flattened into fc3(ReLU). The resulting outputs from the scalar and visual pipeline are concatenated and processed by two additional layers fc4(ReLU) and fc5(ReLU). The parameters of the GMM are generated with the Softmax (fc6), the identity (fc7) and the non-negative ELU (fc8) activation functions respectively.}
	\label{fig:HybridMDN}
\end{figure*}

Our hybrid model architecture relies on visual as well as scalar features.
Visual features are extracted using a simple CNN architecture and concatenated with scalar features obtained by a sequence of fully connected layers in a subsequent step, similar to Bacchiani et. al \cite{Bacchiani2019}.
A depiction of the hybrid model is illustrated in Fig. \ref{fig:HybridMDN}.

\subsubsection{Input}
The visual input consists of 128 x 256 pixels (width x height).
Each pixel represents \SI{1}{\meter} in longitudinal direction,
and \SI{0.1}{\meter} in lateral direction.
The ego vehicle is located at the center of the map.
The longitudinal scale is based on a reasonable distance that is required for planning at urban velocities over multiple time steps (e.g. for $\SI{50}{\kilo\meter\per\hour}\SI{128}{\meter} \approx \SI{9}{\second}$).
The lateral scale is due to the precision required for scenarios with little lateral clearance.

Further, each pixel encodes an integer denoting a semantic class.
Classes are non-drivable areas, static obstacles, dynamic obstacles (i.e. other agents), and lanes.
Each lane is encoded as a separate class allowing the agent to distinguish between them.
The visual input is one frame made up by two maps, a lane map encoding drivable and non-drivable area as well as lanes, and a map with static and dynamic objects.

The scalar input vector includes state information for each agent, (e.g. position, velocity, heading, desired velocity as well as desired lane).

Due to different scales, all inputs are normalized in the range of $[-1, 1]$.

\subsubsection{Output}
In order to generate a valid GMM, the activation functions of the model need to be carefully chosen.
As the mean can be both positive or negative, the identity function for \meanLon and \meanLat is used.
Since the sum of the mixing coefficients need to sum to one,
a softmax activation function for \mixLon and \mixLat is used.
Lastly, the covariance matrix has to be positive semidefinite.
An exponential activation function suggested by Bishop \cite{Bishop1994} can solve this problem and avoid variances close to zero.
However, exponential functions can lead to numerical instability during training. 
Hence, we use the non-negative ELU proposed by \cite{Brando2017} for \varLon and \varLat, see \eqref{eq:nnELU}.

\begin{equation}\label{eq:nnELU}
\text{NNELU} = \text{ELU}(1,x) + 1
\end{equation}

\subsection{Cooperative Planning Algorithm}\label{subsec:mcts}
The cooperative planning algorithm used to generate the data with and integrate the heuristic in as well as benchmark the heuristic against is described in our previous work \cite{Kurzer2018a}.
Based on MCTS the algorithm iteratively improves value estimates for actions.
The basic approach is depicted in Fig. \ref{fig:MCTSSelectionExpansion} and Fig. \ref{fig:MCTSSimulationUpdate}.

In the following we briefly describe the basic concept of MCTS,
that is required to understand how a priori knowledge can be integrated.
However, the interested reader is referred to \cite{Browne2012} or \cite{Kurzer2018a} for a more elaborate description of MCTS or problem specific adaptations respectively.

Monte Carlo methods approximate a quantity using a sampling based approach.
Applied to a Markov decision process (MDP), modeling a sequential decision making problem, sampling can be used to generate different trajectories (i.e. a tuples of states $\statet$ and actions $\actiont$) through the MDP.

The return $\return$ of a trajectory $\trajectory$ is the accumulated reward $\rewardt$ at time step $t$, when taking action $\actiont$ in state $\statet$, see \eqref{eq:Return}.

\begin{equation}\label{eq:Return}
R(\trajectory) = \sum_{(\statet, \actiont) \in \trajectory}{r_t(\statet,\actiont)}
\end{equation}

The Monte Carlo estimate of the action value $Q_\policy(\state, \action)$ is the average of the returns over all trajectories $\trajectory \in \trajectoryspace$ sampled from policy $\policy$, starting in state $\state$ and taking action $\action$, see \eqref{eq:MonteCarloEstimate}.

\begin{equation}\label{eq:MonteCarloEstimate}
\hat{Q}_\policy(s,a) = \frac{1}{|\trajectoryspace|}\sum_{\trajectory \in \trajectoryspace \sim \policy}{R(\trajectory)}
\end{equation}



Based on the current state of the MDP,
MCTS estimates the action value in four distinct phases for each iteration, until a terminal condition is met (e.g. till a computational or time budget is exceeded).
As MCTS is an anytime algorithm \cite{Browne2012}, it will always return an estimate.

\subsubsection{Selection}
During the selection phase the UCT (Upper Confidence Bound for Trees) value, 
see \eqref{eq:UCT}, for a state action tuple is calculated,
and the successor state with the maximum UCT value is selected.
This process repeats itself, until a state is encountered, that has not been fully explored (i.e. not all available actions in the state have been tried), see Fig. \ref{fig:MCTSSelectionExpansion}.

In equation \eqref{eq:UCT}, the first term fosters exploitation of previously explored actions with high action values.
The second term ensures that all actions from a given state are being tried at least once,
with $N(s)$ being the number of times the state $\state$ has been visited and $N(s,a)$ the number of times action $\action$ has been taken in that state.
A constant factor $c$ is used to balance both terms, where larger values for $c$ foster exploration.
\begin{equation}\label{eq:UCT}
UCT(s,a) = Q_\policy(s,a)+c\sqrt{\frac{N(s)}{N(s,a)}}
\end{equation}

\subsubsection{Expansion}
\begin{figure}
	\fns
	\centering
	\def\svgwidth{\columnwidth}
\begingroup%
  \makeatletter%
  \providecommand\color[2][]{%
    \errmessage{(Inkscape) Color is used for the text in Inkscape, but the package 'color.sty' is not loaded}%
    \renewcommand\color[2][]{}%
  }%
  \providecommand\transparent[1]{%
    \errmessage{(Inkscape) Transparency is used (non-zero) for the text in Inkscape, but the package 'transparent.sty' is not loaded}%
    \renewcommand\transparent[1]{}%
  }%
  \providecommand\rotatebox[2]{#2}%
  \newcommand*\fsize{\dimexpr\f@size pt\relax}%
  \newcommand*\lineheight[1]{\fontsize{\fsize}{#1\fsize}\selectfont}%
  \ifx\svgwidth\undefined%
    \setlength{\unitlength}{675.00457764bp}%
    \ifx\svgscale\undefined%
      \relax%
    \else%
      \setlength{\unitlength}{\unitlength * \real{\svgscale}}%
    \fi%
  \else%
    \setlength{\unitlength}{\svgwidth}%
  \fi%
  \global\let\svgwidth\undefined%
  \global\let\svgscale\undefined%
  \makeatother%
  \begin{picture}(1,0.40260815)%
    \lineheight{1}%
    \setlength\tabcolsep{0pt}%
    \put(0.63999521,0.37472054){\color[rgb]{0,0,0}\makebox(0,0)[lt]{\lineheight{0}\smash{\begin{tabular}[t]{l}\fns Expansion\end{tabular}}}}%
    \put(0,0){\includegraphics[width=\unitlength,page=1]{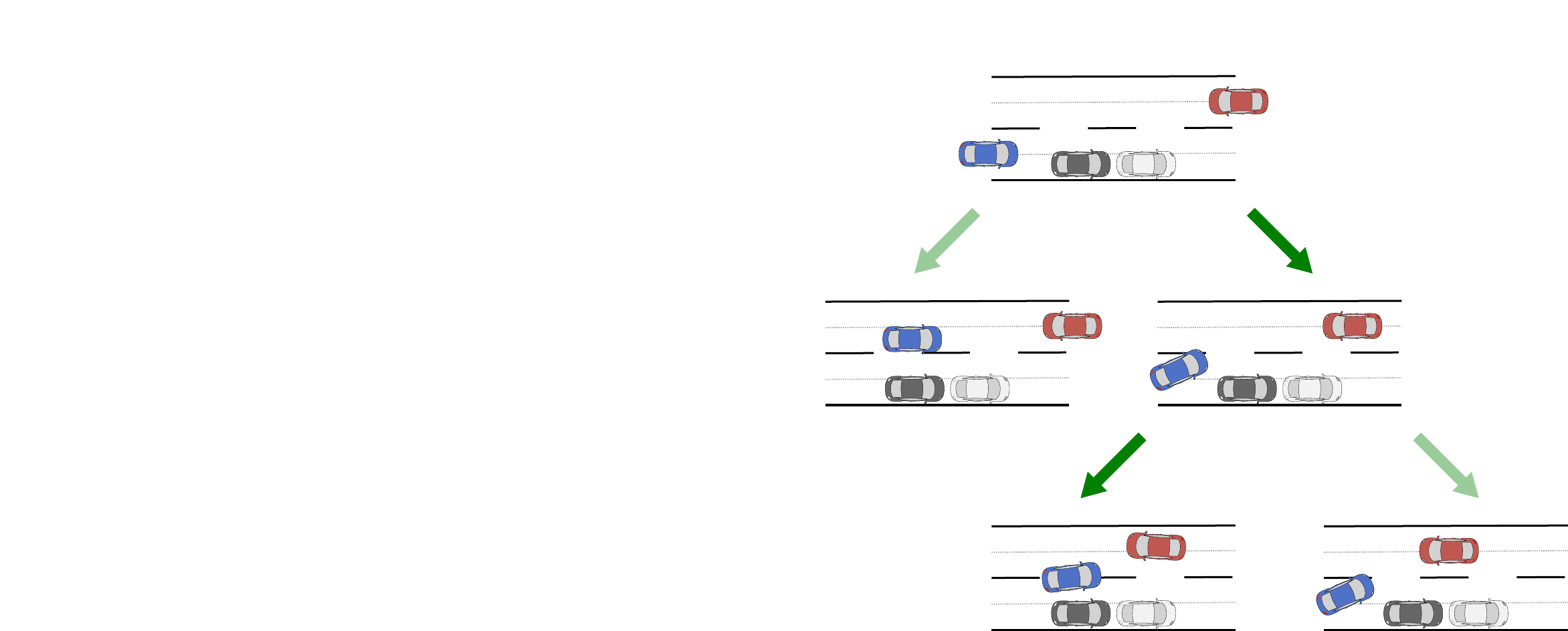}}%
    \put(0.12207931,0.37472054){\color[rgb]{0,0,0}\makebox(0,0)[lt]{\lineheight{0}\smash{\begin{tabular}[t]{l}\fns Selection\end{tabular}}}}%
    \put(0,0){\includegraphics[width=\unitlength,page=2]{mcts_selection_expansion.pdf}}%
  \end{picture}%
\endgroup%

	\caption{Selection and Expansion in MCTS for the scenario depicted in Fig. \ref{fig:MCTSMDN} ; Selection descends the tree by maximizing UCT values until an under-explored state leaf node is found, that gets subsequently expanded with an untried action. Figure adapted from \cite{Kurzer2018a}}
	\label{fig:MCTSSelectionExpansion}
\end{figure}

A state that has untried actions left, gets expanded by sampling an action at random, see \eqref{eq:UniformRandom}, from the action space and executing this action reaching a successor state, see Fig. \ref{fig:MCTSSelectionExpansion}.

\begin{equation}\label{eq:UniformRandom}
a \sim U[\min(\actionspace), \max(\actionspace)]
\end{equation}

\subsubsection{Simulation}
After the expansion a simulation of subsequent random actions is conducted
until a terminal condition is reached (i.e. the planning horizon or an illegal action was sampled resulting in an invalid state), evaluating the quality of the previous expansion, see Fig. \ref{fig:MCTSSimulationUpdate}.

\subsubsection{Backpropagation}
\begin{figure}[h!]
	\fns
	\centering
	\def\svgwidth{\columnwidth}
\begingroup%
  \makeatletter%
  \providecommand\color[2][]{%
    \errmessage{(Inkscape) Color is used for the text in Inkscape, but the package 'color.sty' is not loaded}%
    \renewcommand\color[2][]{}%
  }%
  \providecommand\transparent[1]{%
    \errmessage{(Inkscape) Transparency is used (non-zero) for the text in Inkscape, but the package 'transparent.sty' is not loaded}%
    \renewcommand\transparent[1]{}%
  }%
  \providecommand\rotatebox[2]{#2}%
  \newcommand*\fsize{\dimexpr\f@size pt\relax}%
  \newcommand*\lineheight[1]{\fontsize{\fsize}{#1\fsize}\selectfont}%
  \ifx\svgwidth\undefined%
    \setlength{\unitlength}{675.00457764bp}%
    \ifx\svgscale\undefined%
      \relax%
    \else%
      \setlength{\unitlength}{\unitlength * \real{\svgscale}}%
    \fi%
  \else%
    \setlength{\unitlength}{\svgwidth}%
  \fi%
  \global\let\svgwidth\undefined%
  \global\let\svgscale\undefined%
  \makeatother%
  \begin{picture}(1,0.60305018)%
    \lineheight{1}%
    \setlength\tabcolsep{0pt}%
    \put(0.59777307,0.57516256){\color[rgb]{0,0,0}\makebox(0,0)[lt]{\lineheight{0}\smash{\begin{tabular}[t]{l}\fns Backpropagation\end{tabular}}}}%
    \put(0,0){\includegraphics[width=\unitlength,page=1]{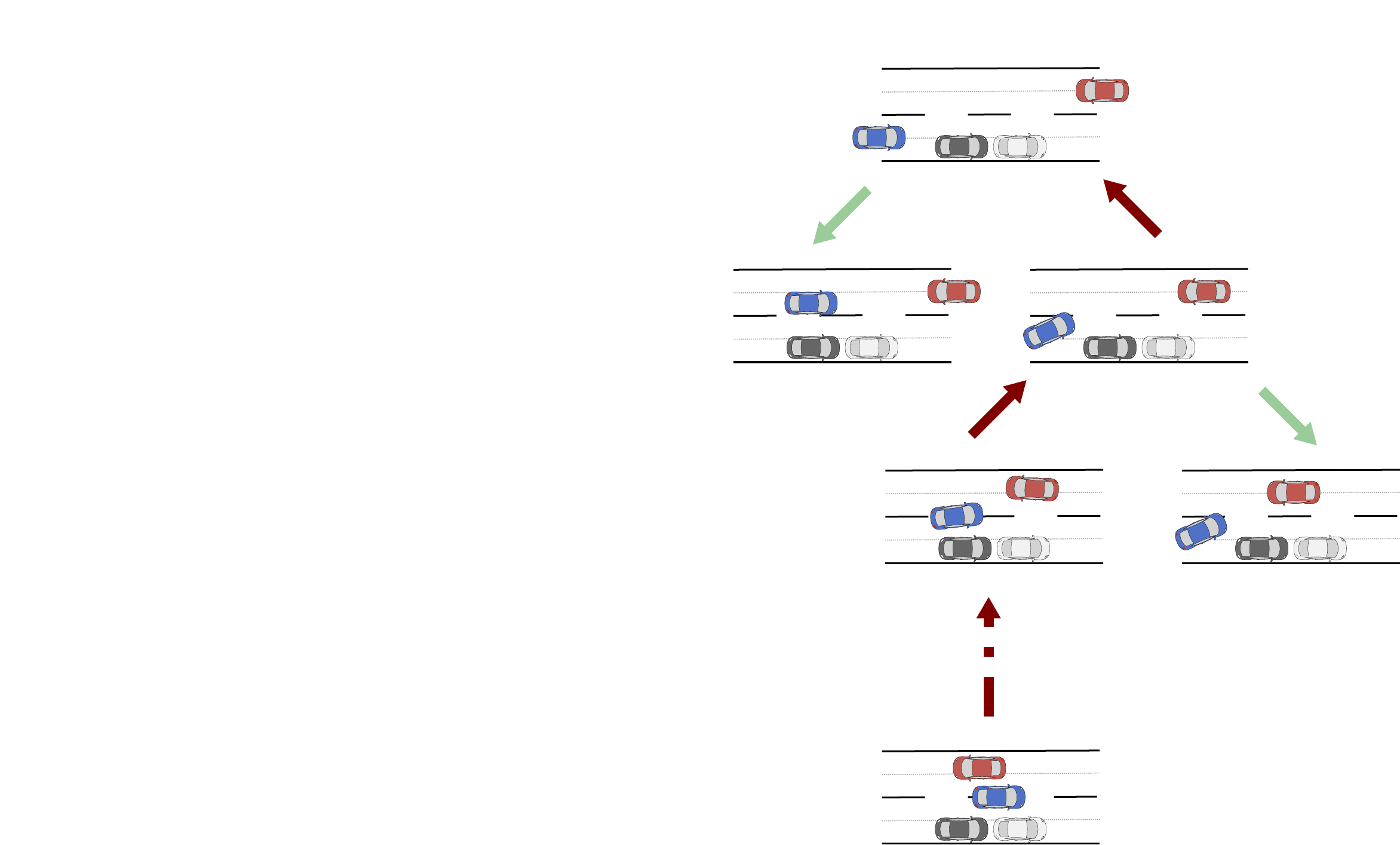}}%
    \put(0.11333269,0.57516256){\color[rgb]{0,0,0}\makebox(0,0)[lt]{\lineheight{0}\smash{\begin{tabular}[t]{l}\fns Simulation\end{tabular}}}}%
    \put(0,0){\includegraphics[width=\unitlength,page=2]{mcts_simulation_update.pdf}}%
  \end{picture}%
\endgroup%

	\caption{Simulation and Backpropagation in MCTS for the scenario depicted in Fig. \ref{fig:MCTSMDN} ; Simulations are run until the end of the planning horizon, after which the result gets backpropagated along the taken trajectory. Figure adapted from \cite{Kurzer2018a}}
	\label{fig:MCTSSimulationUpdate}
\end{figure}

Finally, the return $\return$ of the trajectory generated by the simulation is backpropagated to all states along the trajectory, see Fig. \ref{fig:MCTSSimulationUpdate},
and the action values for all actions of the trajectory are updated, see \eqref{eq:Update}.

\begin{equation}\label{eq:Update}
\hat{Q}_\policy(\state,\action) = \hat{Q}_\policy(\state,\action) + \frac{1}{n}R(\state,\action) - \hat{Q}_\policy(\state,\action)
\end{equation}

Cooperation is ensured by selecting actions based on a combined reward of all agents.
Given enough samples, the algorithm converges to the optimal solution, e.g. the combination of trajectories with the highest cumulated reward.

\subsection{Data Generation and Training}
Due to the absence of a real world cooperative driving data set for urban driving scenarios as well as the cost and time requirements for the creation of such a data set,
we decided to create our own data set using the simulation from our previous work \cite{Kurzer2018a}.

In order to generate a diverse set of training data, we used 15 different scenarios. of which eight scenarios were adapted from Ulbrich et al. \cite{Ulbrich2015}.
A short description and videos of all scenarios can be found online \footnote{\url{http://url.fzi.de/MCTS-MDN-IV}}.

Using the cooperative planning algorithm from \cite{Kurzer2018a},
each scenario was solved 85 times, generating a data set with roughly 1,275,000 expert actions (change in velocity $\Delta v_\text{lon}$ and change in lateral position $\Delta y_\text{lat}$).
At each run the scenarios were initialized at random.
This means, that the position, heading, size, velocity, and desired velocity for each agent and obstacle as well as the width of the road were altered.
Further, we augmented the data by using the point of view of each agent in the scene at every time step, creating n-times the data for n agents and shifted non-ego agents in the scalar input vector $\featuresscalar$.
While this increased the size of the data set, it mainly achieved better generalization,
as the DNN is designed for a fixed number of agents, i.e. eight.
Consequently, for the most part some inputs of the scalar input vector are empty.
To avoid performance degeneration from position one to eight in the vector,
non-ego agents were shifted for each time step through all slots.

Following the data creation, we normalized the number of trajectories for each semantic action class \cite{Kurzer2018a} (i.e. driving straight without velocity change was over represented).

Before being fed into the MDN the input data was further normalized.

Based on the action samples $\action \in O$ of the action space for a given trajectory, two Gaussian mixture models with two and three components were fitted.
The parameters of the GMM constitute the labels for the training.
Using the data set and the corresponding labels, we trained a DNN using the negative-log likelihood for a given sample of the MCTS belonging to the predicted Gaussian mixture model.

We use the TensorFlow/Keras API \cite{tensorflow2015}, to build and train our MDN depicted in Fig. \ref{fig:HybridMDN}.
The training was conducted with a learning rate $l_r = 1\mathrm{e}{-3}$,
in combination with the Adam optimizer,
a batch size of 32 and L2-regularization of the covariance.

As for the loss function, we employ the negative log-likelihood of the observed weighted samples $O$ for each of the $G$ agents in combination with a parameterized L2 cost for the covariance, see \eqref{eq:lossFunction}.
The loss is based on the number of agents $G$ and sampled actions per agent and thus needs to be normalized.
The L2 cost allows tweaking of the resulting probability density function.
\begin{equation}\label{eq:lossFunction}
L = -\frac{1}{G}\sum_{g=1}^{G}{\frac{1}{|O_g|}\sum_{\action \in O_g}{\log\sum_{k=1}^{2}\phi_{gk}\mathcal{N}(\action|\mu_{gk},\Sigma_{gk}) + \alpha\|\Sigma_{gk}\|_2}}
\end{equation}

\subsection{Integration}
With the goal to improve results at a lower computational cost the prior knowledge of the MDN needs to be available in the planning algorithm.
Depending on the exact task and goal, different integration strategies or a combination of them are feasible.

\subsubsection{Expansion Policy}
The integration of prior knowledge in the expansion policy steers the algorithm towards areas of the action space that should yield high returns.
The probability of expanding a specific action $\action$ is proportional to the
value of the MDN for the given features $\features$ and action $\action$, see \eqref{eq:GaussianMixtureSampling}.

\begin{equation}\label{eq:GaussianMixtureSampling}
\action \propto \sum_{k=1}^{K}{\phi_k(\features)\mathcal{N}(\action|\mu_k(\features),\Sigma_k(\features))}
\end{equation}

\subsubsection{Selection Policy}
The integration in the selection policy weighs the exploration term of UCT proportional to the
value of the MDN for the given features $\features$ and $\action$ a, see \eqref{eq:GaussianMixtureSampling}, so actions, lying within high density regions of the MDN are more likely to be selected.

\begin{equation}
UCT(s,a) = Q_\policy(s,a)+p(\action|\features)c\sqrt{\frac{N(s)}{N(s,a)}}
\end{equation}

\subsubsection{Simulation Policy}
An integration in the simulation policy would be identical to the integration in the expansion policy, however, has not been explored in this work, as the computational overhead would be multiple times larger than in the expansion policy.

The inference of the MDN is conducted in C++, requiring less than \SI{5}{\milli\second} for one forward pass. 

\section{Evaluation}\label{sec:evaluation}
\begin{figure}
	\fns
	\begin{subfigure}{\columnwidth}
		\centering
		\def\svgwidth{\columnwidth}
		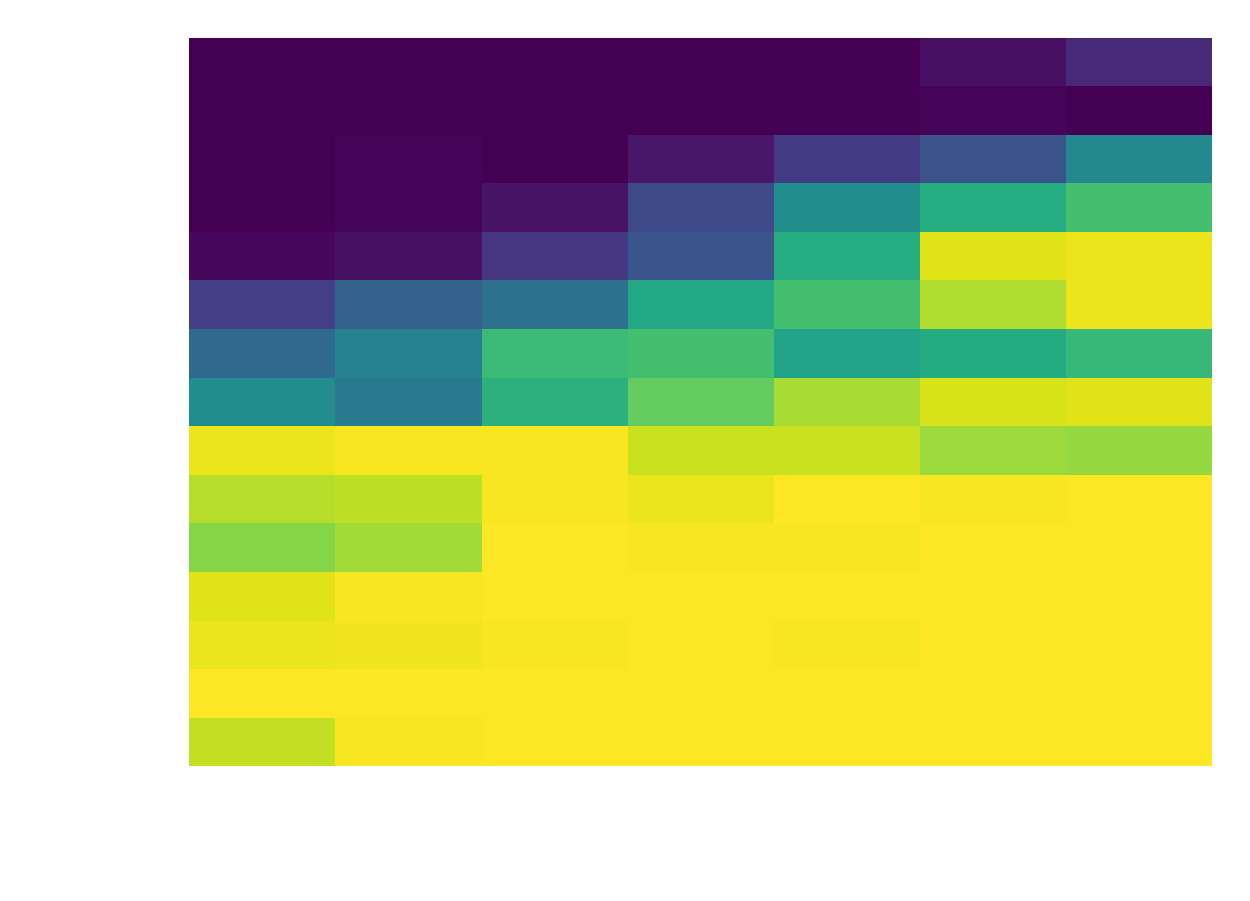
		\caption{MCTS baseline}
		\label{fig:HeatmapMCTS}
	\end{subfigure}
	\begin{subfigure}{\columnwidth}
		\centering
		\def\svgwidth{\columnwidth}
		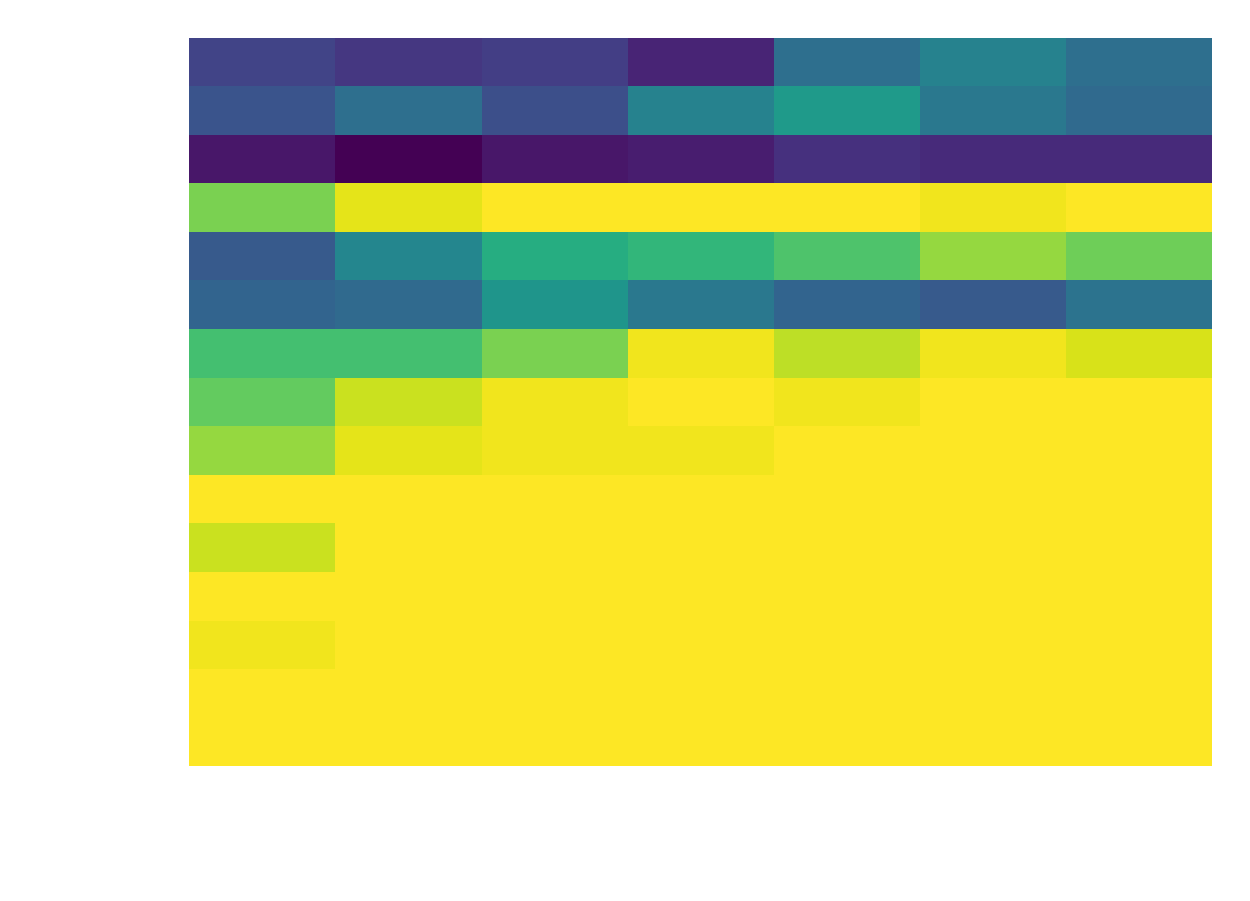
		\caption{MCTS\textsl{}+MDN 2 root w/o selection}
		\label{fig:HeatmapMCTSMDN2}
	\end{subfigure}
	\begin{subfigure}{\columnwidth}
	\centering
	\def\svgwidth{\columnwidth}
	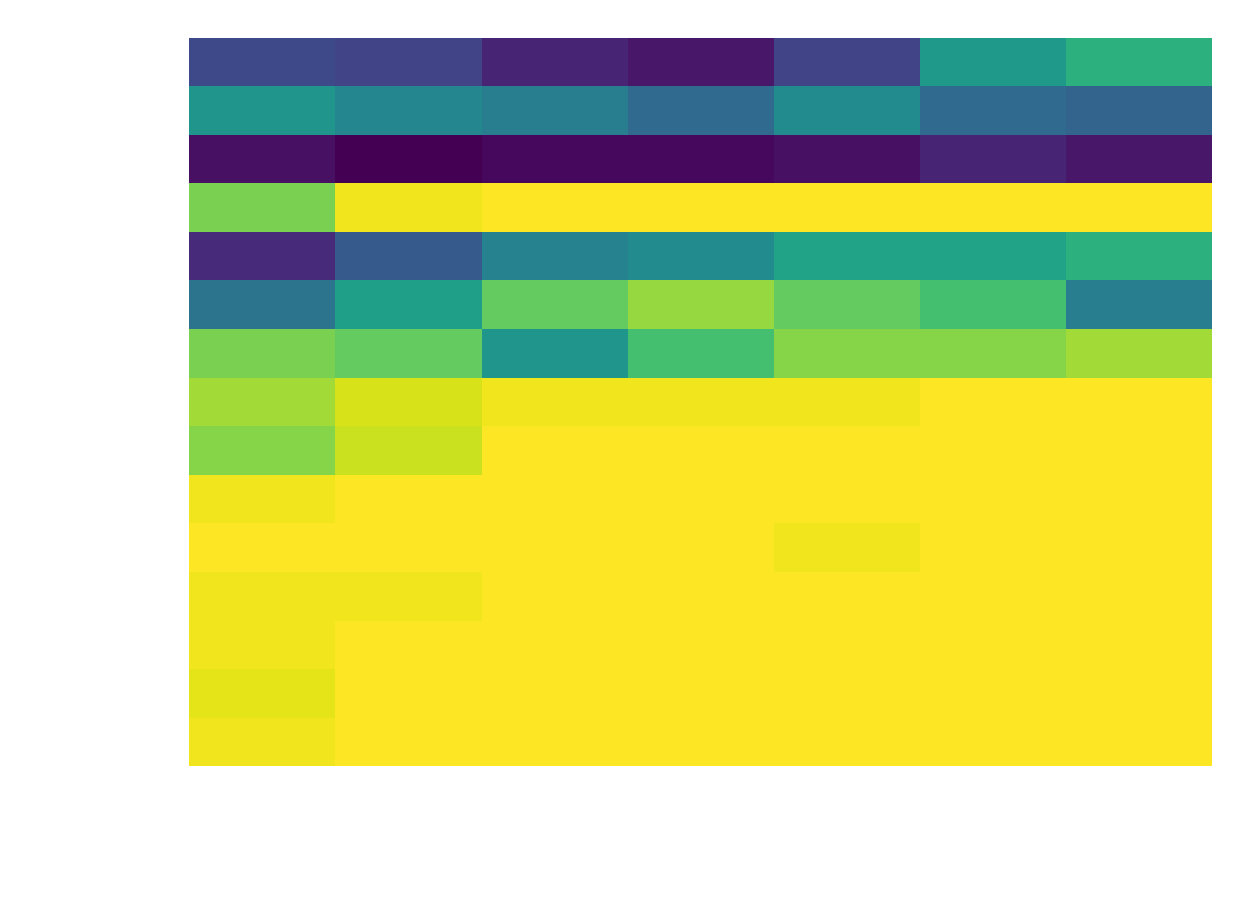
	\caption{MCTS+MDN 3 root w/o selection}
	\label{fig:HeatmapMCTSMDN3}
	\end{subfigure}
	\caption{Evaluation of the success rate (i.e. $1-\text{collision rate}$) for a) the MCTS baseline version and the MCTS+MDN versions with two b) and three c) components used only for the root node expansion without selection for each of the scenarios.}
	\label{fig:Heatmaps}
\end{figure}


\begin{figure}
	\fns
	\begin{subfigure}{\columnwidth}
	\centering
	\def\svgwidth{\columnwidth}
	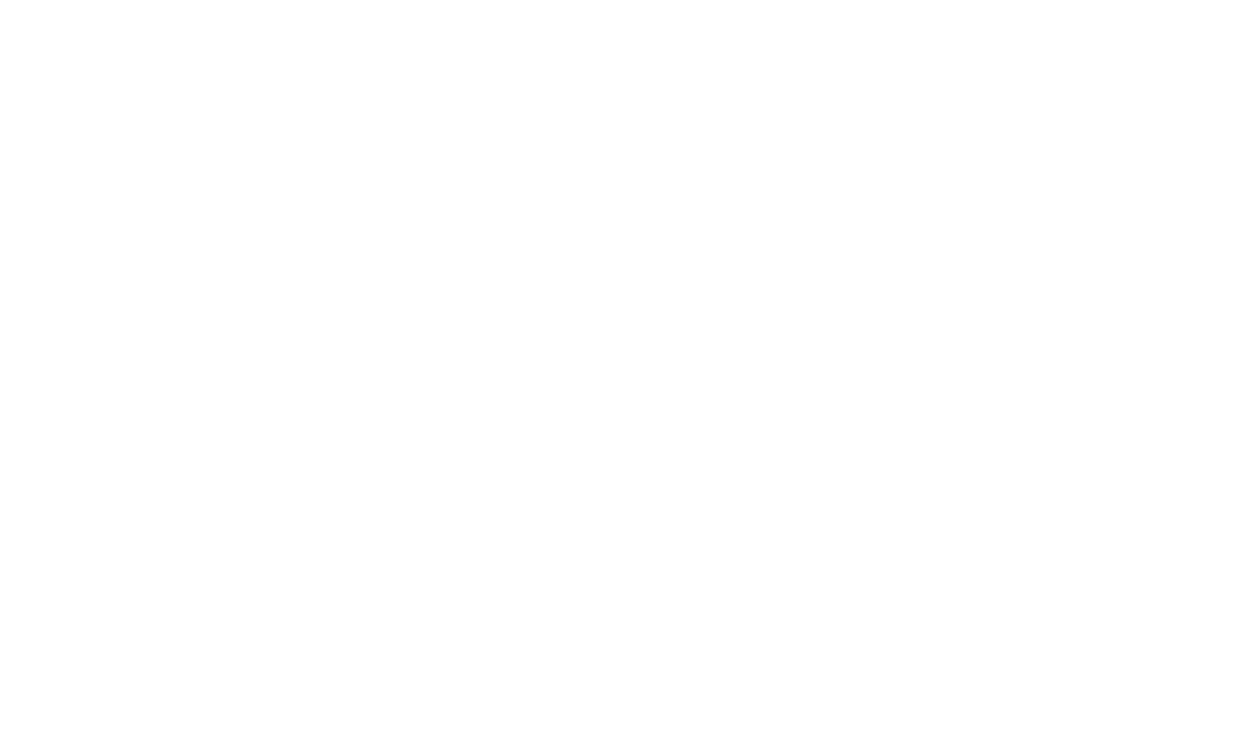
	\caption{Mixture density networks with two and three components used for the root node expansion with and without selection}
	\label{fig:ComparisonSuccessrate}
	\end{subfigure}
	\begin{subfigure}{\columnwidth}
	\centering
	\def\svgwidth{\columnwidth}
	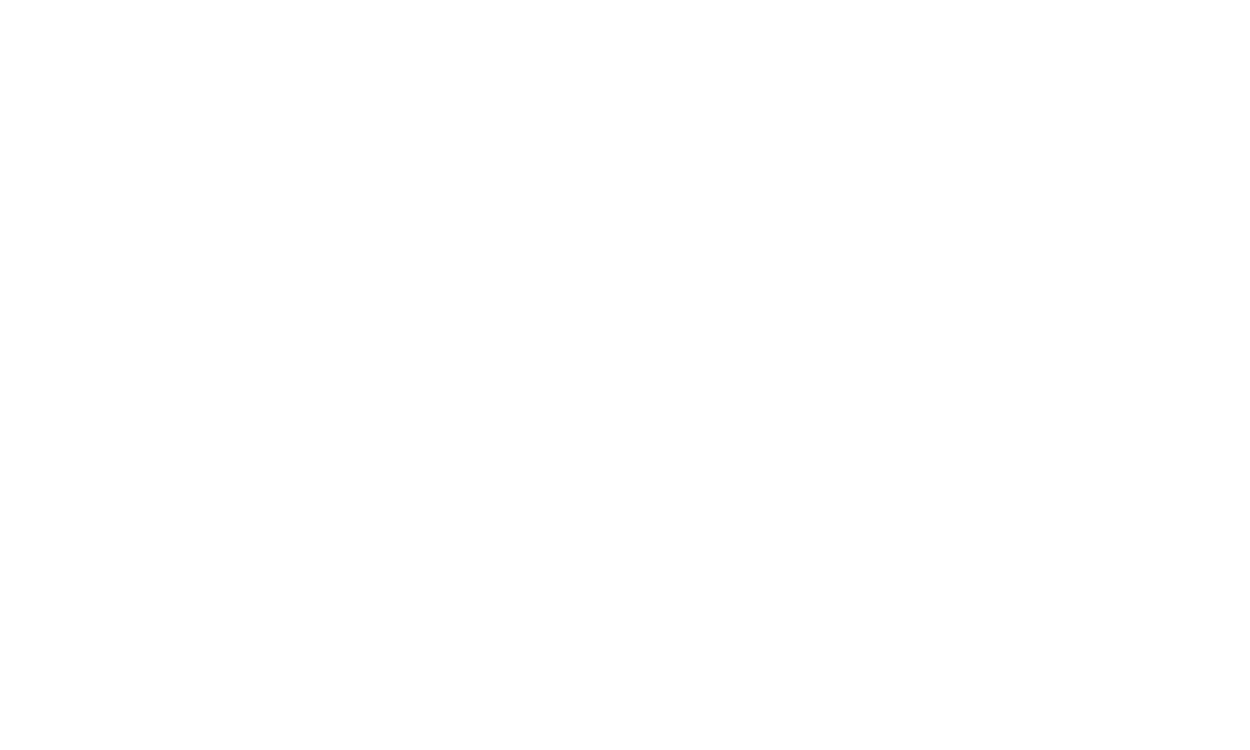
	\caption{Mixture density networks with two and three components used for every expansion with and without selection}
	\end{subfigure}
	\caption{Evaluation of the success rate (i.e. $1-\text{collision rate}$) for the baseline MCTS version (yellow) and the MCTS+MDN versions and integration strategies over all scenarios for different numbers of MCTS iterations }
	\label{fig:ComparisonSuccessrate}
\end{figure}

We evaluated the performance of the baseline MCTS, used to generate the training data with, against the integration of the MDN in the MCTS.
Both the two and three component versions of the MDN were integrated in the MCTS in four different ways:
\paragraph{root w/ selection}
The MDN is used in the expansion policy only in the root node as well as in the selection policy on all nodes.
\paragraph{root w/o selection}
The MDN is used in the expansion policy only in the root node.
\paragraph{all w/ selection}
The MDN is used in the expansion policy on all nodes as well as in the selection policy on all nodes.
\paragraph{all w/o selection}
The MDN is used in the expansion policy on all nodes.

The evaluation of the MDN by itself was conducted by sampling 1,000 different actions from the resulting Gaussian mixture model and choosing the action with the highest probability density at each step in the scenario.

Using each of the proposed solutions the scenarios were each run 50 times, the baseline version 100 times.

As the performance of the MCTS and, thus, also the combination of the MCTS with the MDNs depend
on the number of iterations being run, we ran all with an increasing number of iterations.
Fig. \ref{fig:Heatmaps} depicts the results from the baseline MCTS vs the integration of the prior knowledge provided by the MDNs with two and three components in the MCTS.

Scenarios SC15, SC13, and SC14 were neither solved by the MCTS nor by the combination of MDN and MCTS.
The poor performance on SC14 and SC15 is mainly due to the fact, that these scenarios are especially hard given their obstacle configuration combined with four and eight vehicles respectively.
SC13 on the other hand is usually solved with higher iteration counts and enhancements to the MCTS mentioned in \cite{Kurzer2018a} that are not enabled in the baseline version.
Major improvements were achieved in SC11 and SC12.
While the success rate of SC07 increases slightly for a low number of iterations,
the performance drops sharply once more than 1,000 iterations were used.

Other scenarios behave as expected, generally yielding better results with fewer computational resources.
A summary over the performance averaged over all scenarios is depicted in Fig. \ref{fig:ComparisonSuccessrate}. The absolute average improvement is approximately \SI{18}{\percent} up to 500 iterations (root w/o selection) and vanishes almost completely after 4,000 iterations (w/ selection) and after 8,000 (w/o selection).

While there is no difference in overall performance between the MDN versions with two and three mixture components, three components perform considerably better on SC07. Intuitively this could be due to the three homotopy classes that exist in this scenario, namely, merge before the first, behind the second or between both vehicles.


\section{Conclusions}\label{sec:conclusion}
This paper proposes a method to accelerate multi-agent trajectory planning algorithms for automated vehicles in situations that require cooperation.
Due to the high level representation of the environment, the proposed approach is applicable to
a variety of cases without specific retraining.
While the proposed approach yields a considerable improvement for lower numbers of iterations,
the performance gain decays quickly once 2,000 iterations are exceeded.
The naive integration of the mixture density network requires further tuning and approaches that make a frequent inference computationally feasible.
Additionally, tests with real world data, shall be conducted to reveal whether modifications are needed for deployment on an actual vehicle.


\section{Acknowledgments}
We wish to thank the German Research Foundation (DFG) for funding the project Cooperatively
Interacting Automobiles (CoInCar) within which the research leading to this contribution was conducted. 
The information as well as views presented in this publication are solely the ones expressed by the 
authors.






\bibliographystyle{IEEEtran}
\bibliography{IEEEabrv,04_mendeley-export/library}

\end{document}